# Sparse neural networks with large learning diversity

Vincent Gripon, *Student Member, IEEE* and Claude Berrou, *Fellow, IEEE*
Telecom Bretagne, Electronics department
UMR CNRS Lab-STICC
Brest, France
name.surname@telecom-bretagne.eu

*Abstract*—Coded recurrent neural networks with three levels of sparsity are introduced. The first level is related to the size of messages, much smaller than the number of available neurons. The second one is provided by a particular coding rule, acting as a local constraint in the neural activity. The third one is a characteristic of the low final connection density of the network after the learning phase. Though the proposed network is very simple since it is based on binary neurons and binary connections, it is able to learn a large number of messages and recall them, even in presence of strong erasures. The performance of the network is assessed as a classifier and as an associative memory.

## I. INTRODUCTION

Neural network principles and error correcting coding concepts have sometimes been associated, whether for instance to perform decoding with the help of formal neurons [1] or to use codes to improve learning in perceptrons [2]. However, the question of neural networks offering by construction error correcting ability had not really been addressed until the recent work presented in [3]. In this latter paper, networks based on bipartite weighted graphs linking messages and quasi-orthogonal codes have proved themselves much superior to classical Hopfield neural networks (HNN) [4] in terms of storage diversity. The distinction between diversity (the number of messages that the machine is able to learn) and capacity (the whole learnt information) was a key point in the design of such networks. Contrary to HNN, these networks do not intend to store messages as long as the size of the network and this sparsity characteristic will be kept here. As was said in the conclusion of [3]: "From a cognitive point of view, it is better to learn (and possibly combine) 1000 messages of 10 characters than to learn 10 messages of 1000 characters".

The coded Hopfield networks introduced in [3] face an important drawback due to the necessity of an exhaustive (maximum likelihood) decoding procedure. In recent years, the theory of error correcting codes has considerably evolved with the introduction of distributed coding and iterative decoding. Turbo codes [5] have paved the way for this kind of distributed approach which later found another success with the revival of Low Density Parity Check (LDPC) codes [6] [7]. The new type of neural networks we propose in this paper also uses distributed coding but the coding and decoding rules are quite different from the turbo or LDPC codes: no algebra, no parity check and no extrinsic information. In particular, this latter concept was introduced in [5] in order to minimize correlation effects in message passing decoding. We do not need here such precaution in the decoding process. The messages being stored as geometric patterns, precisely fully interconnected polyhedra, that is cliques, spatial correlation is actually exploited to regenerate them totally from some known edges. For instance, if AB, AC, CD are known in a tetrahedron, then all the six edges (the previous ones plus AD, BC, BD) are known.

The message learning principle, which is then performed through a pattern learning process, relies on a "sparse coding" rule. The expression "sparse coding" is familiar to neurobiologists who use it to express that few neurons among a large population are firing simultaneously at a given time [8]. In the proposed network, this sparse coding rule is pushed to the extreme since only one neuron, in a particular population and in normal conditions, is authorized to be active. This strong sparsity constraint allows the network to learn many messages while keeping its connection density at a low level. Finally, because messages are borne by cliques and since cliques are defined by binary elements (either a vertex or not, either an edge or not), the network is fully binary. This endows it with interesting robustness and resilience properties, unlike the classical HNN in which the messages are very sensitive to the connection weights.

The rest of this paper is organized in four parts. The first one (section II) recalls the main principles and informational properties of HNN that we will consider as a reference to assess later the quality of the proposed scheme. The second one (sections III and IV) introduces the sparse neural network as well as its learning and recalling algorithms. The third part presents the performance of the proposed network as a particular classifier (section V) and as an associative memory (section VI). Finally, in section VII, some comments about the biological plausibility of the model as well as the possible extension of the work are propounded.

## II. HOPFIELD NETWORKS

In the field of associative memory, HNN are the immediate state of the art reference a new model should be compared to. HNN are very simple dynamic models which offer attractive learning capacity. These networks rely on symmetric, weighted, fully interconnected graphs, with the exception that they contain no loop. Such a graph with $n$ nodes is characterized by $\frac{n(n-1)}{2}$ weights, $w_{ij}, 1 \leq i \leq n, 1 \leq j \leq n$ being the weight of the edge linking nodes $i$ and $j$ ($w_{ij} = w_{ji}$). These weights are directly obtained from a set of $M$ learnt binary messages $\{d_i^m\}_{1 \leq m \leq M}$ according to the following formula:

$$w_{ij} = \begin{cases} \sum_{m=1}^{M} d_i^m d_j^m & \text{if } i \neq j \\ 0 & \text{otherwise} \end{cases} \quad (1)$$

Note that $w_{ij}$ may take $P = M + 1$ different values.

The recall of a learnt message, given only part of it, is iteratively computed using the following formula linking the value of each neuron at time $t+1 : v_i^{t+1}$ to its previous value at time $t : v_i^t$:

$$v_i^{t+1} = \begin{cases} 1 & \text{if } \sum_{j=1}^{n} w_{ij} v_j^t \geq 0 \\ -1 & \text{otherwise} \end{cases} \quad (2)$$

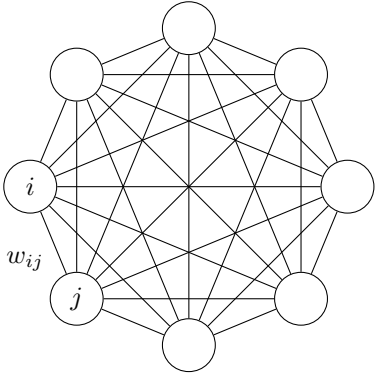

Figure 1. 8 node-Hopfield network model. All nodes are connected to each other through 28 bidirectional edges.

Given a reasonable number of learnt messages, the process converges with a high probability towards the correct one. Noting $log$ the natural logarithm, it has already been demonstrated [9] that this number is upper bounded by:

$$M_{\max} = \frac{n}{2 log(n)} \quad (3)$$

in the case of independent identically distributed random patterns of size $n$ and for $n$ large enough. We call this parameter the diversity of the learning model.

Another important parameter of a memory is its capacity, *i.e.* the maximum amount of data learnt, in bits. As HNN with $n$ neurons learn messages of length $n$, the capacity, that is, the whole amount of possible data learnt, in bits, is given by:

$$C_{\max} = \frac{n^2}{2 \log(n)} \quad (4)$$

A third parameter, the network efficiency $\eta$, which will be of some importance in the sequel, is the ratio between $C_{\max}$ and the amount of information $Q_{\max}$ used by the network when $M = M_{\max}$. A weighted fully connected network having $\frac{n(n-1)}{2}$ connections defined on $P$ levels needs the following amount of memory (in bits) to be specified:

$$Q = \frac{n(n-1)}{2} \log_2(P) \quad (5)$$

For HNN, $P_{\max} = M_{\max} + 1$ and then:

$$Q_{\max} = \frac{n(n-1)}{2} \log_2(M_{\max} + 1) \quad (6)$$

This leads to:

$$\begin{aligned} \eta &= \frac{n}{(n-1)\log(n)\log_2\left(\frac{n}{log(n)} + 1\right)} \\ &\approx \frac{1}{\log(n)\log_2\left(\frac{n}{\log(n)}\right)} \end{aligned} \quad (7)$$

The efficiency of an HNN is low, for instance about $2 \times 10^{-2}$ for $n = 1000$. Moreover, for $n$ tending to infinity, $\eta$ tends to 0, which is not very satisfactory when considering biological plausibility. Several proposals have been made to improve the learning diversity of such networks. For instance, equation (1) above is modified in [10] in such a way that the diversity is increased up to a value $\frac{n}{\sqrt{2\log(n)}}$ but the number of levels $P$ required to express the weights is also increased so much so that the efficiency is not actually improved. Another well-known example of recurrent neural networks with ameliorated diversity is given by the Boltzmann machine [11]. The efficiency of such networks depends a lot on the learning rule parameters. This being an iterative procedure with many steps, the number of levels $P$ is generally too large to have good repercussions on $\eta$, compared to classical HNN. So we will keep the latter as the state-of-the art as far as efficiency is concerned.

In this paper, we are interested in values of $\eta$ close to one. Note that the efficiency, as we defined it, can be actually larger than one because the messages stored in the networks are not arranged in order, as they are in address-indexed memories. Storing $M$ binary messages of length $n$ in a content addressable memory does not theoretically require as much storage space as a single ordered message of length $Mn$.

Now consider again the general expression of $Q$ given by (5). Ideally, this amount of available memory allows the storage of $\frac{n-1}{2} \log_2(P)$ ordered messages of length $n$. If the length of messages is restricted to a value $k < n$, their maximal number becomes $\frac{n(n-1)}{2k} \log_2(P)$. This states that, given a restricted size $k$, the number of messages a network can learn grows quadratically with its size $n$ (whereas linearly when considering messages of length $n$). Obviously, one may point out that the capacity is the same in both cases, but this simple demonstration shows how interesting the learning of short messages can be, in terms of diversity.

However, HNN are not directly adapted to the learning of short-length messages. An original way to exploit this sparsity benefit on data representation is described in the following sections. Surprisingly, the new kind of networks that we will introduce, based on this first sparsity principle, will offer large gains not only in diversity but also in capacity and consequently in efficiency.

## III. CLIQUES AS CODEWORDS OF A GOOD ERROR CORRECTING CODE

In an undirected graph, a clique is a set of nodes such that each one is connected to the others. First, let us observe that an HNN is itself a clique with $n$ nodes and that this clique contains $2^n - n - 1$ cliques of all sizes between 2 and $n$. In this ensemble, consider the subset composed of the $\binom{n}{c}$ $c$-cliques, that is, cliques with $c$ vertices. We can define the minimum Hamming distance $d_{\min}$ between elements of this subset as the minimum number of edges that differ between two cliques. $d_{\min}$ is obtained in the case of only one vertex being different:

$$d_{\min} = 2(c-1) \quad (8)$$

The coding rate $R$ is the ratio between the minimum number of edges necessary to specify a particular clique: $\lfloor \frac{c+1}{2} \rfloor$ and the total number of edges: $\frac{c(c-1)}{2}$. This gives, for $c$ even:

$$R = \frac{1}{c-1} \quad (9)$$

The merit factor $F = R d_{\min}$ then equals 2. This value being greater than 1 means that a clique can be regarded as a codeword of a good error correcting code (by comparison, the well-known (8, 4, 4) Hamming code has also a merit factor equal to 2). The clique, which has then an inherent discrimination capability and which also is of highly biological plausibility [12], will be our elementary informational brick in the construction of new neural networks.

The number of $c$-cliques in a graph with $n$ nodes being potentially very high, these offer theoretically a plentiful memory space for the learning of short-length messages. However, when trying to store messages as cliques in a complete graph without any precaution, it quickly happens that many "false" (non learnt) cliques of any size, resting on the edges of the "true" (learnt) cliques, multiply. In order to avoid this secondary effect, we propose to organize the network in such a way that only cliques with a given size $c$ can be learnt and recalled and that the number of false cliques is kept low. In order to achieve this, as detailed in the next section, the network is partitioned into clusters and within each cluster, a local sparse coding rule is introduced.

## IV. NEURAL NETWORKS WITH LARGE LEARNING DIVERSITY

Let us consider a network with $n$ binary neurons of values $v_i \in \{0;1\}$ and binary edge weights $w_{ij} \in \{0;1\}$. This network is split into $c$ clusters of size $l = n/c$. The fact that clusters are all of the same size offers no other interest but to simplify the equations. Neurons are called fanals, as the learning process is such that only one of them can be activated at a time in its "dark cluster". Moreover, let us set $l$ to a power of 2, such that each fanal of a given cluster may be mapped one to one to a binary vector of length $\kappa = log_2(l)$. Let us denote by $f : \{-1;1\}^\kappa \to [|1;l|]$ this mapping, where $[|1;l|]$ is the subset of integers between 1 and $l$. Therefore, with an input binary message $m$ of length $k = c\kappa$ is associated a unique set of fanal neurons (one per cluster) using this simple transformation: $C : m = \underbrace{m_1 m_2 \ldots m_c}_{\text{each of size } \kappa \text{ bits}} \to (f(m_1), f(m_2), \ldots, f(m_c))$.

This network relies on a multipartite graph, *i.e.* there is no connection between neurons within a same cluster. As the process described before is totally (and easily) reversible, learning a message $m$ of $k$ bits is equivalent to learning the corresponding pattern $C(m)$.

Therefore, to learn a given message $m$ of $k$ bits, the network learns the pattern $C(m)$. Projected onto the network, this pattern is represented by a completely off network except for the fanal neurons materializing the pattern. These few fanals are then fully interconnected (if a connection already exists, then it remains unchanged), leading to the corresponding clique. The expression "neural clique" is also familiar to neurobiologists to describe such groupings, though not so precisely formalized [12]. Figure 2 illustrates this learning process.

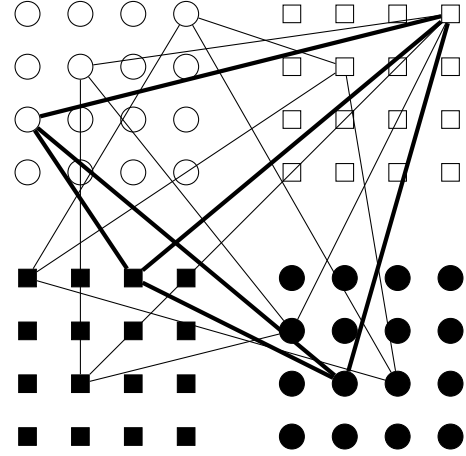

Figure 2. Learning process illustration. The pattern to learn (with thick edges) connects fanal neurons from 4 clusters of 16 fanals each (filled circles, filled rectangles, rectangles and circles). It represents a geometric figure (in this case a tetrahedron) which is printed in the network.

As by definition these neurons belong to distinct clusters, this construction is compatible with the multipartite structure of the associated graph. Formally, after the learning of $M$ messages $m_1 \ldots m_M$, the weight $w_{(c_1 l_1)(c_2 l_2)}$ linking the $l_1$-th fanal of cluster $c_1$ to the $l_2$-th fanal of cluster $c_2$ is set to:

$$w_{(c_1 l_1)(c_2 l_2)} = \begin{cases} 1 & \text{if } c_1 \neq c_2 \text{ and } \begin{cases} \exists m \in \{m_1 \ldots m_M\}, \\ C(m)_{c_1} = l_1 \\ C(m)_{c_2} = l_2 \end{cases} \\ 0 & \text{otherwise} \end{cases} \quad (10)$$

This result is independent of the order in which messages are presented. Moreover, learning a new message can be done at any moment, with no need to normalize.

Referring to the previous section, with $P = 2$ and as the maximum number of connections in such a network is $\frac{(c-1)n^2}{2c}$, the number of learnable ordered messages of length $k$ is upper bounded by:

$$M'_{\max} = \frac{(c-1)n^2}{2c^2 log_2\left(\frac{n}{c}\right)} \quad (11)$$

For instance, with $c = 4$ clusters, a network made of $n = 2048$ neurons maps to a bound of $4.4 \times 10^4$ possible ordered messages. This bound increases to $5.7 \times 10^5$ for $n = 8192$ (typically the size of a neocortical column). Once again, let us point out that this bound does not apply to non ordered messages whose number can exceed this value.

Obviously, a first importance parameter to assure a good categorization between learnt and not learnt messages is the network density. A density close to 1 leads to an overloaded network, which cannot retrieve learnt messages. After the learning of $M$ uniformly distributed random messages, connections in the network may be considered independent, which is not much of an approximation if $M \gg c$. The expected density $d$ is then directly connected to $M$ by the following formula:

$$d = 1 - \left(1 - \frac{1}{l^2}\right)^M \approx \frac{M}{l^2} \text{ when } M \ll l^2 \quad (12)$$

Figure 3 shows the evolution of the density $d$ with the number $M$ of random messages learnt, for four values of $l$. Note that this density does not depend on the total number of neurons $n$ neither on the number of clusters $c$ but on the cluster size $l$, meaning that from this point of view, and given a total number of neurons, a small number of clusters is preferable. The choice of $c$ may also depend on other criteria, such as the targeted retrieval error rate (see section VI and relation (30)).

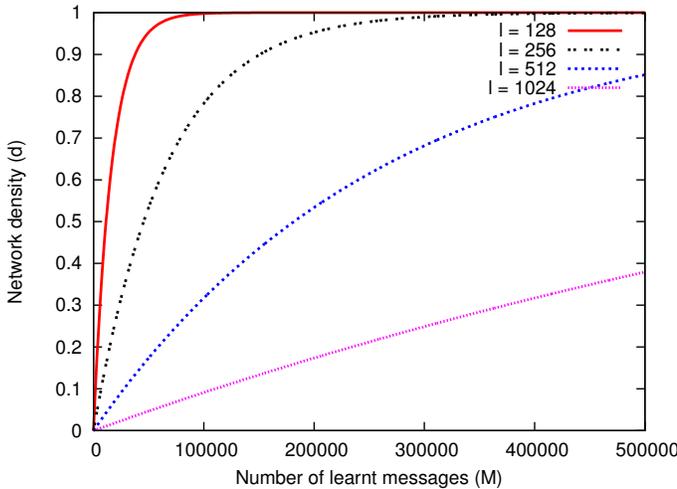

Figure 3. Evolution of the network density when learning random messages for 4 cluster sizes.

Note that the organization of the network in clusters has reduced the number of possible connections with respect to the complete graph ($\frac{(c-1)n^2}{2c}$ to be compared with $\frac{n(n-1)}{2}$) but this reduction, and therefore the cut in available memory, are low and acceptable (25% for $c = 4$, for instance).

So, each message is encoded by the activation of a relatively small set of neurons. Actually, the sparse coding rule which underlies the activity of neurons inside each cluster is not a novel one. In the theory of error correcting coding, this is known as a constant-weight code [13]. We can define such a code by three parameters: the length, the weight (i.e. the number of 1 in each codeword) and the overlapping (i.e. the maximal number of 1 that codewords can share at the same location). These parameters are respectively equal to $l$, 1 and 0 in our case, leading to the sparsest possible code we could adopt. This code is a weak one since its minimum Hamming distance is only 2, but it benefits from a very simple decoding modality which consists in choosing the codeword in which the 1 is at the place where the magnitude of symbols is the highest, regardless of the type of additive noise.

Therefore, the kind of neural network we propose can be seen as a concatenation of weak codes, in the same way as LDPC codes, for instance, whose local codes have also a minimum Hamming distance of 2. The global error correcting power of concatenated (or distributed) code is generally much higher than that of local codes.

The next section details the associated decoding algorithm.

### A. Local decoding

The local decoding process must take good decisions from its particular point of view. Beneath is presented a local decoder using a very simple neural circuit where the neuron model is described in Figure 4. This local decoder has already been presented in [3], and is reintroduced using a new formalism.

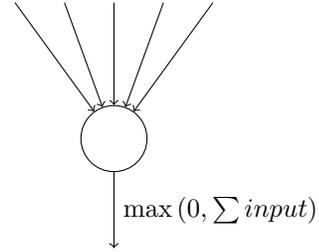

Figure 4. Neuronal model used for local decoding. The neuron routine is split into two steps: first it sums the different inputs, then it keeps the obtained sum if it is higher than 0.

A hard output maximum likelihood (ML) decoder maps an input $in$ to an admissible output $out$ with the maximum *a posteriori* probability. That is to say that the output produced by an ML decoder is:

$$out = \arg\max_{a \in A} P(a|in) \quad (13)$$

where $A$ is the set of admissible outputs and $a$ one among them. Actually, the decoder presented hereafter is a soft ML decoder, meaning that if several admissible outputs share the maximum *a posteriori* probability, the output of the decoder will be partially erased since there is no reason to make a choice. For instance, if two such admissible outputs are the binary words 001 and 011 then the soft ML decoding output should be $0X1$ where $X$ denotes an erased character. The next paragraphs present a way to build a soft ML decoder using the neuronal model of Figure 4.

Let us consider a set of admissible outputs $A \subset \{-1; 1\}^\kappa$. With each of these outputs $a \in A$ is associated a unique neuron $N(a) \in N_a$ ($\#N_a = \#A$ where $\#A$ denotes the cardinal of $A$). These neurons are also called fanals (as only one of them should be activated in good cases). The fanals are connected to the input neurons $I_1 \ldots I_\kappa$ through binary connections. The bipartite graph produced is complete, meaning that the two possible values for a connection are $-1$ or $1$ (not $0$). Let us denote by $g_{ij}$ the connection between input neuron $I_i$ and fanal $N(a_j)$. This network is fully characterized by $A$ through:

$$g_{ij} = (a_j)_i \quad (14)$$

The value of a neuron $N(a)$ is given by the function $v : N(a) \to v(N(a))$. Figure 5 depicts an example of such a decoder.

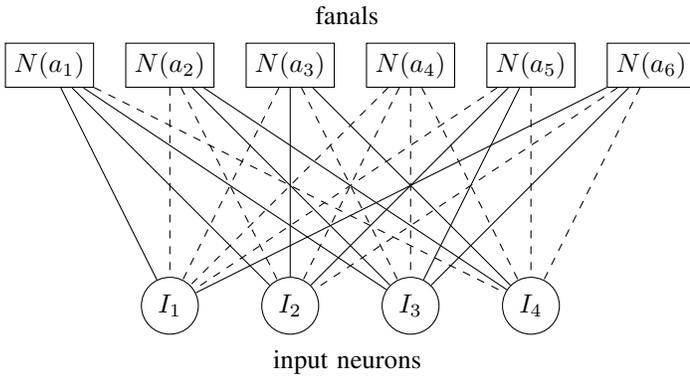

Figure 5. Example of a neural decoder for a code of $\#A = 6$ codewords of length $\kappa = 4$ ($a_1$ =+1+1+1-1, $a_2$ =-1-1+1+1, $a_3$ =-1+1-1+1, $a_4$ =-1-1-1-1, $a_5$ =-1+1+1-1 and $a_6$ =+1-1+1-1). Dashed edges map to $-1$ weights.

The way each cluster elects its fanal relies on a very simple rule: the more signals received, the more likely they are to survive. Value 1 is then assigned to the survivor(s) while the other neurons are switched off (value 0).

The local decoding process is then described by the following algorithm:

$$\forall j \leq \#A, v(N(a_j)) \leftarrow \sum_{i=1}^{\kappa} g_{ij} v(I_i) \quad (15)$$

$$v_{Max} \leftarrow \max_{j \leq \#A} v(N(a_j)) \quad (16)$$

$$\forall j \leq \#A, v(N(a_j)) \leftarrow \begin{cases} 1 & \text{if } v(N(a_j)) = v_{Max} \\ & \text{and } v_{Max} \geq \sigma \\ 0 & \text{otherwise} \end{cases} \quad (17)$$

$$\forall i \in [|1; \kappa|], v(I_i) \leftarrow \begin{cases} 1 & \text{if } \sum_{j \leq \#A} g_{ij} v(N(a_j)) \geq 0 \\ -1 & \text{otherwise} \end{cases} \quad (18)$$

Let us observe that (17) allows several fanal neurons to switch on simultaneously. This property allows this decoder to produce a soft output. Step (17) uses a threshold $\sigma$ to control the activity of this local decoder if needed. This threshold depends on the application: in classification the largest possible value leads to the best results whereas in associative memories it has to be more carefully balanced.

Nevertheless, one should object that this algorithm is fully neural except step (16) which supposes that the neural network is able to find the maximum activation value among a population of neurons. The following paragraphs give an example of a neural structure that can solve this problem.

To find the maximum value of activated neurons, we simply extend the following identity, which holds for any real $x$ and $y$:

$$\max(x, y) = \frac{x+y}{2} + \left|\frac{x-y}{2}\right| \quad (19)$$

Figure 6 depicts a representation of a neural adaptation of this equation. This representation requires one of the input numbers to be nonnegative.

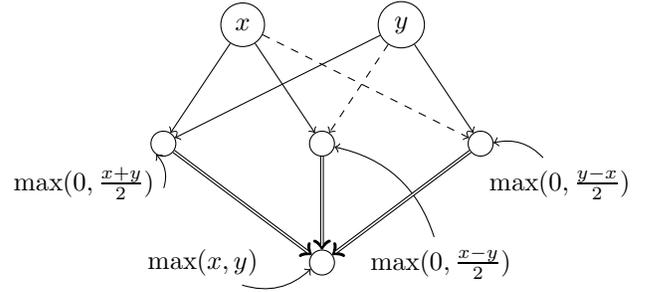

Figure 6. Maximum selector using neurons. Solid line edges have weight $\frac{1}{2}$, dashed lines weight $-\frac{1}{2}$, and double lines 1.

The maximum of any set of numbers may then be neurally computed using a cascade of this operator. This construction is depicted in Figure 7. The hypothesis is more loose as just one input must be positive or null (and not one by sub-adder).

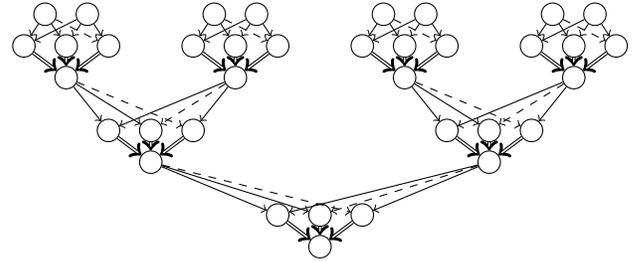

Figure 7. Maximum selector over $2^q$ (here: $q = 2$) values with neurons. Solid line edges have weight $\frac{1}{2}$, dashed lines weight $-\frac{1}{2}$, and double lines weight 1.

From this construction, finding the maximum activity among a population of neurons can be achieved using a reasonable number of added neurons. However, its complex architecture has no biological plausibility. The aim of this paper is not to investigate the plausible models for the local decoders. One should find interesting literature on the subject, considering Kohonen maps [14] for instance. More generally, a biological system of the type "winner-take-all" seems quite plausible as far as minimum energy issues are considered.

Finally, Figure 8 shows a full network using this construction with $c = 4$ clusters.

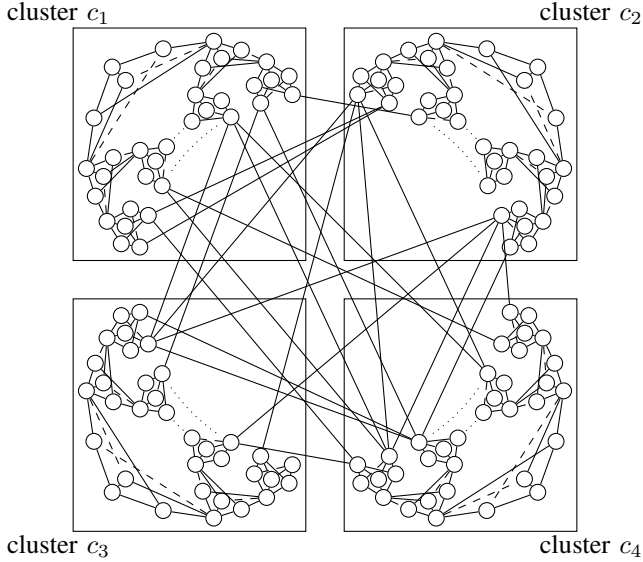

Figure 8. Representation of a full network using simple neural local decoders. This network is composed of $c = 4$ clusters.

## B. Global decoding

This section describes how to retrieve the correct pattern given the decisions of local decoders.

Let us denote by $n_{ij}$ ($i \leq c, j \leq l$) the $j$-th fanal of the $i$-th cluster and extend the function $v$ to these neurons. Then the global decoding is performed using the following iterative algorithm:

$$\forall i, j, v(n_{ij}) \leftarrow \sum_{i'=1}^{c} \sum_{j'=1}^{l} w_{(i'j')(ij)} v(n_{i'j'}) + \gamma v(n_{ij}) \quad (20)$$

$$\forall i, \text{ use the local decoding on cluster } i \quad (21)$$

(20) formalizes a message passing algorithm including a memory effect, with parameter $\gamma$. This memory effect is necessary to achieve good performance. In particular, it assures that a learnt message, if presented as input, will always be recognized. On the other hand, its value must not be too high to allow error correction. From now on, this parameter will be considered to be equal to 1, its minimum non zero value since all neuron values are integers. One may note that this is very similar to authorizing loops in the graph but in addition to the fact that the memory effect allows a better control on the process, it also offers better biological plausibility than loops.

When considering inputs with a lot of noise or erasures, this algorithm can become iterative. However, a single iteration is enough in many cases.

Figure 9 depicts the clustered network decoding principle with a simple example. This network has learnt the word "brain" by creating a clique. If the word is then presented to the network with an erased first character, the 4 connections coming from the other characters will contribute to retrieving the missing "b". To achieve this, all the fanals of the first cluster will be scanned and the one obtaining the highest score from the signals sent by the activated "r", "a", "i" and "n" from the other clusters, will be elected. Of course, if the network has also learnt "train", there will be a conflict between "b" and "t" and the two corresponding fanals will be activated, leading to ambiguity. Moreover, if the network has learnt "grade" and "gamin" in addition to "brain", it will recognize "grain" because the connexions between "g" and the last four letters of "brain" have been established. This latter case is typical of a "false" (non learnt) clique whose probability of appearance decreases with the increase of $c$.

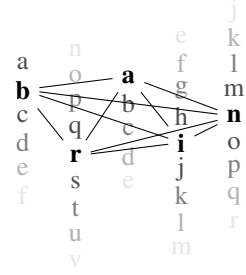

Figure 9. Example of a clustered network having learnt the word "brain".

Note that no care is taken about correlation in the message passing decoding. This point marks a strong difference with the way distributed error correcting decoders perform. In particular, the clique-decoder does not calculate any extrinsic piece of information as necessarily used in turbo or LDPC decoders. Indeed, when trying to reconstruct a clique from a part of it, correlation (whose effects are propagated and maintained within cycles) is more a help than a handicap. On the biological level, one can imagine that life developed its informational strategy by exploiting correlation and not combatting it.

The number of iterations in the repeated computation of (20) and (21) depends both on the application and on the integrity of the incoming data possibly affected by noise or erasures. This network can achieve unprecedented performance when used both as a classification model and as an associative memory. The next sections describe these two applications.

## V. CLASSIFICATION

Neural networks have for long invested the field of classification. One objective of classification is to generalize learning sets, as in pattern recognition. Other branches of classification do not require this extrapolation as the learning sets are exhaustive. In this case, the objective is to decide whether a given message is part of a learnt class or not, or more specifically whether it has been learnt or not. Possible application fields are intrusion detection systems, set implementations, etc. Only this second aspect of classification is considered in this paper: to accept a given input message if and only if it has been learnt. The set of messages accepted by a network is the set $F$ of messages that remain unchanged after a given number of iterations.

Let us denote by $E$ the set of learnt messages. Therefore, a measure of performance over the network is a measure of differences between sets $E$ and $F$. (20) implies $E \subseteq F$, so the probability that a learnt message is not recognized, called the first kind error probability, is zero. Hence the only possible errors remaining are of second kind: unlearnt messages accepted by the network.

Two different measures are described beneath. The first one is the measure $P(x \in F | x \notin E)$, i.e. the error probability of second kind. As $E$ is supposed to be of reasonable size ($E \ll 2^k$), this measure is close to $P(x \in F)$. Given a threshold $\sigma = c$ coupled with $\gamma = 1$, one can easily estimate this probability in case of a single iteration. Indeed, a given input on the network will remain activated if all the connections between its associated fanals exist. Considering again that connections are independent, as was the hypothesis to establish (12), this leads to the following formula:

$$P(x \in F) \approx d^{\frac{c(c-1)}{2}} \qquad (22)$$

Note that this is also the probability that $c$ neurons randomly chosen, one in each cluster, are forming a clique in the network.

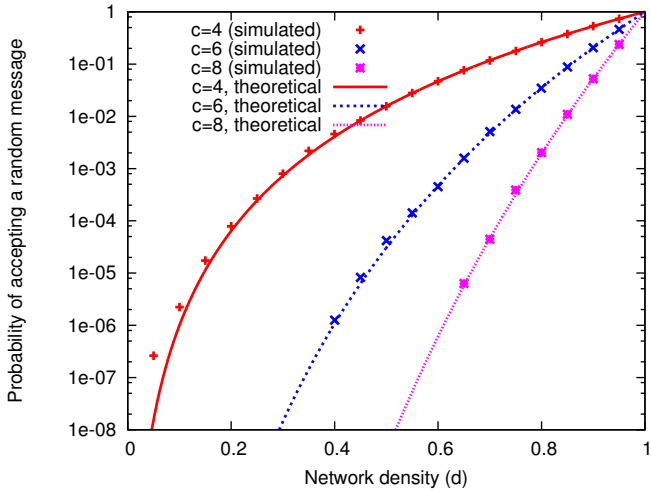

Figure 10. Probability to accept a random message for 3 numbers of clusters with size $l = 512$ and for $\sigma = c$. Both simulated points and theoretical curves are represented.

Figure 10 shows the evolution of this probability for three numbers of clusters with size $l = 512$ and for $\sigma = c$. Note that this probability remains acceptable even with a very high density (up to $80\%$ for $c = 8$). Coupled with Figure 3, it shows that a network with $n = 4096$ neurons and $c = 8$ clusters may learn more than $400000$ messages of length $72$ with still very good discrimination ability with non learnt messages.

However, one must understand that even if the error probability is low, there may be a lot of unlearnt messages that could be accepted in an exhaustive test. The number of accepted messages can be estimated from (22) this way:

$$\#F = 2^k P(x \in F) \approx 2^k d^{\frac{c(c-1)}{2}} \qquad (23)$$

So, another interesting measure is to compare the size of $F$ with that of $E$. Figure 11 draws the evolution of the ratio $\frac{\#(F-E)}{\#E}$ for three numbers of clusters with two sizes: $l = 256$ and $l = 512$. This figure shows that this ratio is very good for a reasonable density. One may note that increasing the size of clusters does not improve performance as the number of possibly accepted messages is considerably enlarged. For instance, changing $l = 256$ into $l = 512$ increases this size by a factor of $256$ in the case of $c = 8$. Meanwhile, the number of learnt messages has only grown by a factor of four.

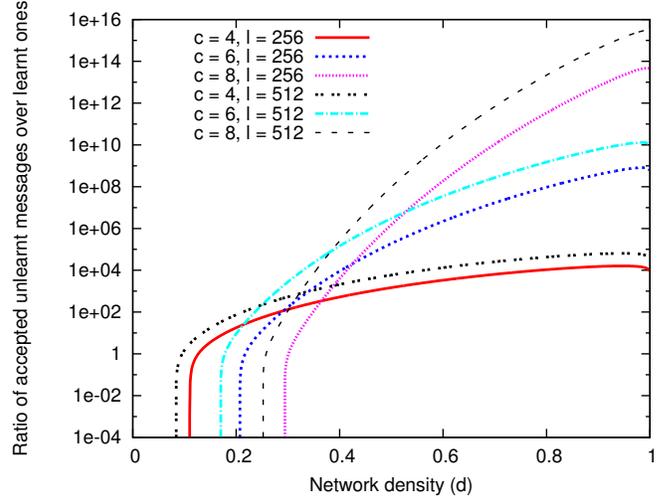

Figure 11. Ratio of the number of unlearnt accepted messages over the learnt ones, in an exhaustive go/no-go test, in function of the network density and for 3 different numbers of clusters of size $l = 256$ and $l = 512$.

Table I shows the comparison of performance of HNN and of the proposed network for $c = 4$ and $l = 512$, with the same amount of memory used. The material needed for the local decoders is not taken into account as it does not depend on the learning sets. Those results show the very good performance of the proposed network to achieve a go/no-go sort. Note also that the efficiency of the proposed network, in this experiment, exceeds $100\%$. Taking into account the previous remarks about $\eta$ (sections II and IV), this result should not surprise.

| Model | HNN | Proposed network | ratio |
|---|---|---|---|
| Memory used (bits) | $1.6 \times 10^6$ | $1.6 \times 10^6$ | 1 |
| $n$ | 740 | 2048 | $\times 2.8$ |
| Message length | 740 | 36 | $\div 21$ |
| First kind error probability | $9\%$ | $0\%$ | $\div \infty$ |
| Second kind error probability | almost 0 | almost 0 | $\approx 1$ |
| Diversity | 56 | 60000 | $\times 1071$ |
| Capacity | $4.1 \times 10^4$ | $2.2 \times 10^6$ | $\times 52$ |
| Efficiency | $2.6\%$ | $137\%$ | $\times 52$ |

Table I
COMPARISON OF PERFORMANCE BETWEEN THE HNN AND THE PROPOSED NETWORK WITH $c = 4$ AND $l = 512$ FOR THE SAME AMOUNT OF MEMORY USED, IN THE CASE OF A GO/NO-GO SORT APPLICATION.

## VI. ASSOCIATIVE MEMORY

Another interesting aspect of the proposed network is its ability to retrieve data given only part of it. This is possible as long as the set of learnt random messages ($M$) is small compared to possible ones ($2^k$), giving a reasonable typical distance between learnt messages[1]. Contrary to the classification problem, first kind error probability is no longer zero as clusters with no provided information may induce an ambiguous decision on others. We will now consider the threshold $\sigma$ to be zero.

Once again, it is possible to estimate the error probabilities considering a single iteration in the network. However, unlike the classification problem, iterations can significantly increase performance. So, after a single iteration and when only one cluster is not provided with information, the probability of electing the correct erased fanal is given by the following equation:

$$P_{retrieve} = \left(1 - d^{c-1}\right)^{l-1} \quad (24)$$

and the probability that no ambiguity is produced in the other clusters is:

$$P_{remain} = \begin{cases} \left(\left(1 - d^{c-2}\right)^{l-1}\right)^{c-1} & \text{if} \quad \gamma = 0 \\ 1 & \text{otherwise} \end{cases} \quad (25)$$

Considering that the memory effect is actually used ($\gamma > 0$), the error probability of recovering the message is:

$$P_e = 1 - P_{retrieve} = 1 - \left(1 - d^{c-1}\right)^{l-1} \quad (26)$$

Given (12), this probability can be writen as:

$$P_e = 1 - \left(1 - \left[1 - \left(1 - \frac{1}{l^2}\right)^M\right]^{c-1}\right)^{l-1} \quad (27)$$

More generally, if the number of clusters $c_e$ without provided information is larger than one, this probability becomes:

$$P_e = 1 - \left(1 - \left[1 - \left(1 - \frac{1}{l^2}\right)^M\right]^{c-c_e}\right)^{(l-1)c_e} \quad (28)$$

When the number of messages tends to zero, and for a reasonable cluster size: $l \gg 1$, this probability is close to:

$$P_e \approx l c_e \left[\frac{M}{l^2}\right]^{c-c_e} \quad (29)$$

Figure 12 draws the evolution of the message retrieval error rate when one of the four clusters of size $512$ is not provided with information, in function of the number of learnt messages. This figure also draws the theoretical curve from (28).

The optimal number of clusters, given a targeted error probability $P_0$, a number of neurons $n$ and a proportion of clusters with no input information of $\frac{1}{2}$, can be easily obtained from (29) as follows:

$$c_{opt} = log\left(\frac{n}{2P_0}\right) \quad (30)$$

For instance, the approximated optimal number of clusters for a targeted error probability $P_0 = 0.25$ with $n = 2048$ neurons is 8. Figure 13 draws the evolution of the retrieval error rate when half the clusters of such a network with $c = 8$ have no information, in function of the number of learnt messages and after four iterations. The theoretical curve for a single iteration from (28) is also drawn, showing the interest of iterative process in this situation. The simulation shows that such a network of $2048$ neurons can learn up to $15000$ messages of $64$ bits each and retrieve them with a very high probability even when they are erased up to a half. This is, to our knowledge, unprecedented performance.

[1] However, note that a short distance between messages does not necessarily imply a short spatial distance between associated patterns.

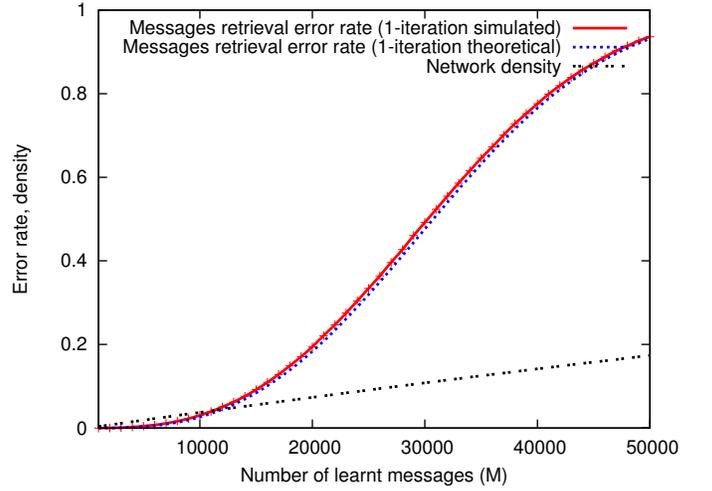

Figure 12. Evolution of the error rate when retrieving a learnt message with 1 cluster with no provided information in a network composed of 4 clusters of size $512$ in function of the number of learnt messages. The simulated and theoretical curves as well as the network density are represented.

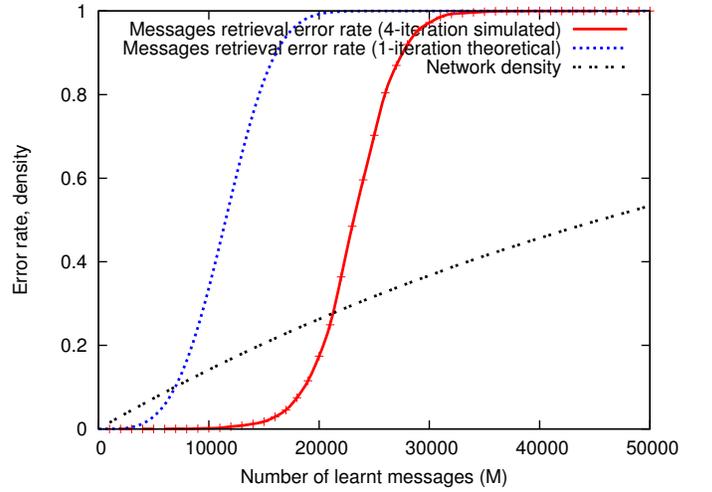

Figure 13. Evolution of the error rate when retrieving a learnt message after 4 iterations with 4 clusters having no information in a network composed of 8 clusters of size $256$, in function of the number of learnt messages. The theoretical curve for a single iteration and the network density are also drawn.

Once again, the performance obtained with the proposed network is dramatically better than that obtained by HNN. Table II compares the two models for the same amount of memory used and half the input erased. One can observe that diversity has been increased a lot compared to the equivalent HNN. This is easily explained as learnt messages are shorter in our model. What is more surprising is the considerable gain in capacity.

| Model | HNN | Proposed network | ratio |
|---|---|---|---|
| Memory used (bits) | $1.8 \times 10^6$ | $1.8 \times 10^6$ | 1 |
| $n$ | 790 | 2048 | $\times 2.6$ |
| Message length | 790 | 64 | $\div 12$ |
| Error probability | 9% | 2% | $\div 4.5$ |
| Diversity | 60 | 15000 | $\times 250$ |
| Capacity | $4.7 \times 10^4$ | $9.6 \times 10^5$ | $\times 20$ |
| Efficiency | 2.6% | 52% | $\times 20$ |

Table II
COMPARISON OF PERFORMANCE BETWEEN THE HNN AND THE PROPOSED MODEL WITH $c = 8$ AND $l = 256$ FOR THE SAME AMOUNT OF MEMORY USED, WHEN BOTH ARE USED AS ASSOCIATIVE MEMORIES.

Figure 14 depicts the gain of capacity from the HNN to the proposed network. The given curve considers a network with $c = 8$ clusters where one is not provided with information and the error probability in the retrieving process is close to $10^{-2}$. This latter condition is severe as equivalent HNN present worse error probabilities even without erased inputs. The figure includes also the theoretical curve for a hypothetic network with efficiency equal to one. The proximity between this latter curve and that of the proposed network is worthy of note.

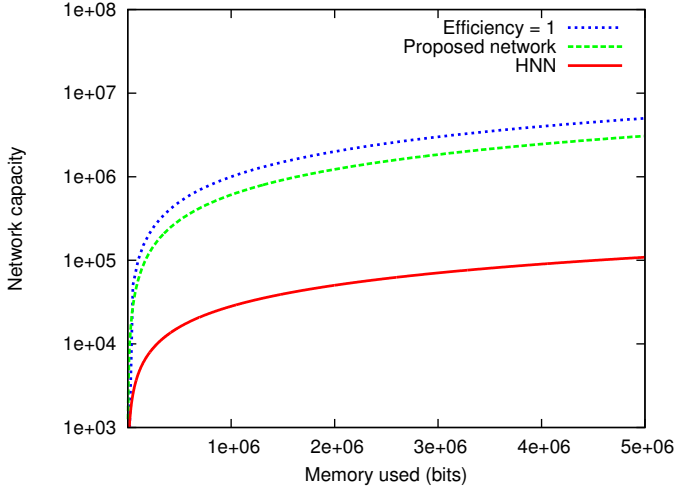

Figure 14. Capacity of the proposed network compared with that of HNN in function of the amount of memory used, in the case of $c = 8$ clusters and a targetted error probability of 0.01 when one is not provided with information. The theoretical curve corresponding to efficiency equal to 1 is also provided.

## VII. CONCLUSION AND OPENING

The ideas and results that were presented in this paper highlight the interest of sparsity in neural networks when judiciously combined with the concepts of distributed error correcting coding and decoding (though not very classical in our case). Sparsity is exploited at three levels: the length of messages, the local constraint on neurons activity and the connection density, the latter being a consequence of the former. To achieve this, messages are materialized by cliques whose discrimination properties are fully exploited in the clustered network. The number of messages that the network is able to learn and recall definitely breaks the sub-linear law of HNN given by (3).

Regarding biological plausibility, the proposed neural network has several assets. First of all, its type of activity is in full concordance with what neurobiologists call "sparse coding". As a network of clusters with local constraints (no more than one active neuron in a cluster, at a time), it can be seen as a grid of "small worlds" [15] with dense local connectivity (for the local decoding) and sparse global connectivity (for the communication between clusters). The connections are binary and therefore resilient. It does not matter a lot if connections have weights 1, 0.9 or 1.1; the decoding process will be successful all the same. The learning process is incremental: up to a certain density level, to a new message learnt correspond a few new connections, without any alteration of the pre-existing messages. Of course and in the same manner as in HNN, bidirectional connections are not biologically acceptable, but to consider them systematically as two-wire cables. Nevertheless, though this possibility was not detailed in the paper, it is conceivable to build comparable networks with directed edges. For instance and on average, a 7-clique with 21 unidirectional edges has as many "informative connections" per node (*i.e.* the average incident degree) as a 4-clique with 6 bidirectional edges. The way to design networks with directed connections, that are more biologically plausible, is then to use slightly more complex geometric patterns.

As for the local decoding modalities presented in section IV.A., the circuit of Figure 7 is far too well structured to have a cellular equivalent. Physically speaking, the research and selection of the neuron having maximum activity concomitantly with the extinction of the others may be interpreted as an energy limitation problem. Perhaps also, but this is merely speculative, glial cells may play a role in this selection process.

Be that as it may, the values of $l$ (some hundreds to some thousands) we have considered in the presentation of this new type of neural network are comparable to the number of neurons in neocortical columns and this gives a high degree of biological plausibility to this network family. By combining many of these networks in a manner yet to be defined, one can imagine being able to obtain machines with very interesting properties. For instance, because learnt cliques share vertices and/or edges, one can contemplate the possibility of designing machines able to learn a lot of messages and to produce new information by association, fusion, or crossbreeding.

ACKNOWLEDGEMENTS

The authors would like to thank Jean-Claude Carlach (Orange Labs) for the rich exchanges of views they often had together about distributed coding and the anonymous reviewers for their relevant comments and suggestions on the contents and organization of this paper.